\documentclass{article}
\usepackage[T1]{fontenc}
\usepackage{spconf,amsmath,amssymb,graphicx,hyperref}
\usepackage{booktabs,multirow,array}
\usepackage[table]{xcolor}
\usepackage{tikz}
\usetikzlibrary{arrows.meta,positioning,calc,shapes.geometric}
\usepackage{enumitem}
\usepackage[capitalise,noabbrev]{cleveref}
\newcommand{\method}{\textsc{Tern}}
\newcommand{\bx}{\mathbf{x}}
\newcommand{\bq}{\mathbf{q}}
\newcommand{\bk}{\mathbf{k}}
\newcommand{\bv}{\mathbf{v}}
\newcommand{\be}{\mathbf{e}}
\newcommand{\bo}{\mathbf{o}}
\newcommand{\bu}{\mathbf{u}}
\newcommand{\bz}{\mathbf{z}}
\newcommand{\bS}{\mathbf{S}}
\newcommand{\bI}{\mathbf{I}}
\newcommand{\balpha}{\boldsymbol{\alpha}}
\newcommand{\bphi}{\boldsymbol{\phi}}
\newcommand{\Diag}{\operatorname{Diag}}
\newcommand{\Rb}{\mathbb{R}}
\newcommand{\best}[1]{\textbf{#1}}
\newcommand{\second}[1]{\underline{#1}}
\newcommand{\up}{$\uparrow$}
\newcommand{\dn}{$\downarrow$}

\newcommand{\para}[1]{\smallskip\noindent\textbf{#1}\ }

\title{TERN: A DELTA-RULE MEMORY WITH A SEASONAL REFERENCE AND ONLINE ADAPTATION FOR EPIDEMIC FORECASTING}
\twoauthors{Shunya Nagashima\sthanks{Equal contribution.}}{Neurogica Inc., Japan}{Yuta Funayama$^{\ast}$}{LTS, Inc., Japan}
\begin{document}
\ninept
\maketitle
\begin{abstract}
Weekly influenza surveillance counts guide vaccine distribution and public-health alerts, yet they are hard to
forecast. Each region offers only a few seasons, waves shift in timing and height every year, and information that
helps while a wave grows misleads after its peak, whereas last season's shape stays informative for a year. Existing
epidemic graph models and general forecasters read a short fixed window and treat all past information alike, so
they neither exploit earlier seasons nor discard stale associations when the epidemic phase changes. To address
these limitations, we propose \method{}, a forecaster built around a delta-rule fast-weight memory that decays
channel-wise and erases along a learned address under gates driven by local epidemic-phase features, combined with
an explicit seasonal reference and online adaptation. On three Cola-GNN influenza benchmarks, \method{} outperformed
epidemic graph models and general forecasters, matched or exceeded seasonal references, and a controlled comparison
confirmed the contribution of the memory itself.

\end{abstract}
\section{Introduction}
\label{sec:intro}
Weekly influenza surveillance counts are the operational input of public-health forecasting hubs~\cite{mathis2024},
and forecasts several weeks ahead guide vaccine distribution and public alerts. The series are
short and strongly non-stationary. Each region provides a handful of seasons, and the waves change their timing and
height every year. A seasonal outbreak is moreover a sequence of regimes.
Information that was useful during the growth phase of a wave becomes misleading after the peak, whereas the shape
of the previous season stays useful for a year. A forecaster therefore has to forget selectively and to keep
several time scales at once.

Two communities have addressed this setting in isolation. Epidemic-specific deep learning models couple regions through
learned graphs and attention~\cite{colagnn,epignn} or inject mechanistic structure~\cite{einns,camul}, and the
general time-series community has produced increasingly strong forecasters based on patching, inverted attention
and multiscale mixing~\cite{patchtst,itransformer,timemixer,dlinear}. Both read a short fixed window, neither
controls explicitly what is forgotten and what is kept, and the two are rarely compared under one protocol. Recent linear-attention layers provide this control, from the
delta rule~\cite{deltanet} and gated decay~\cite{gdn,kda} to variants that separate erasing from writing~\cite{gdn2,eda}.

To address these limitations, we propose \method{} (Temporal Erase-then-delta Recurrent Network), a forecaster built around an
erase-then-delta memory mixer with channel-wise decay and learned time constants, whose erase and write gates and
erase address depend on local phase features, so that what is erased depends on the epidemic phase. An explicit
seasonal reference and online adaptation complete the model. Our code and configurations are publicly
available.\footnote{\url{https://github.com/Neurogica/TERN}} Our contributions are as follows:
\begin{itemize}[leftmargin=*, nosep]
\item We propose, to our knowledge, the first delta-rule linear-attention forecaster for epidemic surveillance
series, from which Kimi Delta Attention and Gated DeltaNet-2 each differ by a single term.
\item We isolate the contribution of the memory with a controlled comparison that holds context length, seasonal
reference and online adaptation fixed, in which softmax attention degrades RMSE on all three datasets.
\item We present the first side-by-side evaluation of epidemic graph models, general forecasters, seasonal
references and recent zero-shot foundation models (Chronos-2 and TimesFM-3) under one protocol, in which \method{}
is the best uncontaminated model on all three influenza datasets.
\end{itemize}

\section{Related Work}
\label{sec:related}
\para{Deep learning for epidemic forecasting.}
Cola-GNN~\cite{colagnn} introduced the influenza benchmarks used here and learns cross-location attention on top
of an RNN. EpiGNN~\cite{epignn} adds transmission-risk encodings and a region-aware graph learner. Mechanistic
hybrids such as EINNs~\cite{einns} regularise a neural forecaster with compartmental dynamics, and
CAMul~\cite{camul} fuses several data views. Many spatial models do not outperform naive baselines in recent benchmarks~\cite{spatialepibench,epicastbench}.

\para{General forecasting and delta-rule attention.}
Patching~\cite{patchtst}, inverted attention~\cite{itransformer}, multiscale
mixing~\cite{timemixer}, 2D variation modelling~\cite{timesnet}, linear models~\cite{dlinear} and
decomposition-based state-space models with input-dependent time scales~\cite{decompssm} define the state of the
art on long-term forecasting benchmarks, and we re-run the first five under the epidemic protocol.
DeltaNet~\cite{deltanet} writes with the delta rule, Gated DeltaNet~\cite{gdn} adds a scalar forget gate, Kimi
Delta Attention~\cite{kda} makes the decay channel-wise, and Gated DeltaNet-2~\cite{gdn2} and Erase-then-Delta
Attention~\cite{eda} decouple erasing from writing, the latter with an erase step that is addressed independently of the write key.

\definecolor{cBlueB}{RGB}{78,98,170}   \definecolor{cBlueF}{RGB}{198,208,242}
\definecolor{cGreyB}{RGB}{100,100,100} \definecolor{cGreyF}{RGB}{226,226,226}
\usetikzlibrary{backgrounds,fit}
\pgfdeclarelayer{stack}\pgfsetlayers{background,stack,main}
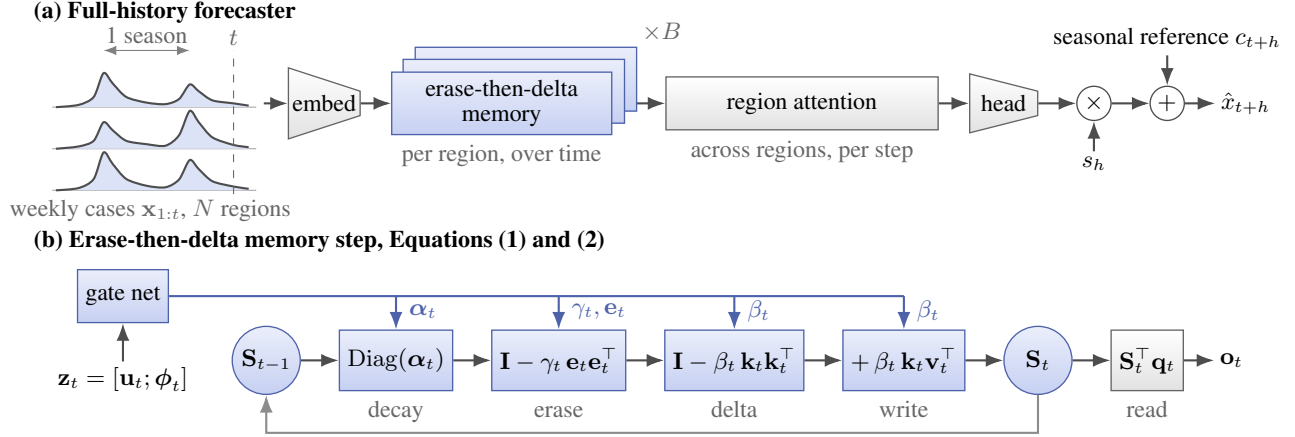
\begin{figure*}[t]
\centering
\begin{tikzpicture}[
  x=1cm, y=1cm, font=\small, >=Latex,
  every node/.style={inner sep=3pt, align=center},
  grey/.style={draw=cGreyB, line width=0.5pt, top color=white, bottom color=cGreyF, minimum height=7mm},
  mem/.style={draw=cBlueB, line width=0.5pt, top color=cBlueF!35!white, bottom color=cBlueF, minimum height=7mm},
  memback/.style={draw=cBlueB, line width=0.5pt, top color=cBlueF!35!white, bottom color=cBlueF, inner sep=0pt},
  enc/.style={draw=cGreyB, line width=0.5pt, trapezium, trapezium angle=72, shape border rotate=270, trapezium stretches=true,
              top color=white, bottom color=cGreyF, minimum height=9mm, minimum width=6mm, inner sep=1.5pt},
  op/.style={draw=cBlueB, line width=0.5pt, top color=cBlueF!35!white, bottom color=cBlueF, minimum height=8mm, inner sep=3pt},
  state/.style={circle, draw=cBlueB, line width=0.5pt, top color=cBlueF!35!white, bottom color=cBlueF, minimum size=9mm, inner sep=1pt},
  sumnode/.style={circle, draw=black!70, line width=0.5pt, fill=white, inner sep=1.2pt},
  var/.style={inner sep=2pt},
  arr/.style={->, thick, black!70},
  barr/.style={->, thick, cBlueB},
  glbl/.style={font=\small, text=cBlueB, inner sep=1.5pt},
  lbl/.style={font=\small\bfseries, inner sep=0pt},
  note/.style={font=\small, text=black!60, inner sep=1pt}
]
\node[lbl, anchor=west] at (0,3.8) {(a) Full-history forecaster};
\begin{scope}[shift={(0.25,1.45)}]
\foreach \r/\a/\b in {2/0.45/0.30, 1/0.35/0.50, 0/0.50/0.40} {
  \begin{scope}[shift={(0,\r*0.55)}]
  \draw[black!35] (0,0) -- (2.7,0);
  \fill[cBlueF!60!white, draw=none] plot[smooth, tension=0.5] coordinates {(0.05,0)(0.35,0.03)(0.55,0.15*\a/0.5)(0.68,\a)(0.8,0.7*\a)(0.98,0.3*\a)(1.25,0.06)(1.5,0.04)(1.68,0.18*\b/0.5)(1.82,\b)(1.94,0.7*\b)(2.12,0.3*\b)(2.4,0.05)(2.6,0.02)} -- (2.6,0) -- (0.05,0) -- cycle;
  \draw[thick, black!70] plot[smooth, tension=0.5] coordinates {(0.05,0)(0.35,0.03)(0.55,0.15*\a/0.5)(0.68,\a)(0.8,0.7*\a)(0.98,0.3*\a)(1.25,0.06)(1.5,0.04)(1.68,0.18*\b/0.5)(1.82,\b)(1.94,0.7*\b)(2.12,0.3*\b)(2.4,0.05)(2.6,0.02)};
  \end{scope}}
\draw[dashed, black!60] (2.4,-0.05) -- (2.4,1.85); \node[note] at (2.4,2.02) {$t$};
\draw[<->, black!50, thin] (0.68,1.85) -- node[above, note, inner sep=1pt, yshift=1.5pt] {1 season} (1.82,1.85);
\node[note] at (1.3,-0.27) {weekly cases $\bx_{1:t}$, $N$ regions};
\end{scope}
\node[enc] (emb) at (3.85,2.6) {embed};
\node[mem, text width=2.7cm, right=4mm of emb] (blk) {erase-then-delta\\[-1pt]memory};
\begin{pgfonlayer}{stack}
  \node[memback, fit=(blk), xshift=3.2mm, yshift=3.2mm] {};
  \node[memback, fit=(blk), xshift=1.6mm, yshift=1.6mm] {};
\end{pgfonlayer}
\node[note, anchor=south west, inner sep=2pt] at ($(blk.north east)+(3.2mm,3.2mm)$) {$\times B$};
\node[grey, minimum width=3.6cm, right=7mm of blk] (ra) {region attention};
\node[enc, right=4mm of ra] (head) {head};
\node[sumnode, right=5mm of head] (gain) {$\times$};
\node[var, below=4mm of gain] (sh) {$s_h$};
\node[sumnode, right=5mm of gain] (sum) {$+$};
\node[var, right=4mm of sum] (out) {$\hat x_{t+h}$};
\node[var, above=4mm of sum] (ref) {seasonal reference $c_{t+h}$};
\node[note, below=0.9mm of blk] {per region, over time}; \node[note, below=0.9mm of ra] {across regions, per step};
\draw[arr] (3.05,2.6) -- (emb); \draw[arr] (emb) -- (blk); \draw[arr] (ra) -- (head);
\begin{scope}[on background layer] \draw[arr] (blk) -- (ra); \end{scope} 
\draw[arr] (head) -- (gain); \draw[arr] (sh) -- (gain); \draw[arr] (gain) -- (sum); \draw[arr] (ref) -- (sum); \draw[arr] (sum) -- (out);
\node[lbl, anchor=west] at (0,0.78) {(b) Erase-then-delta memory step, \cref{eq:erase,eq:delta}};
\node[var, anchor=west] (z) at (0.25,-1.05) {$\bz_t=[\bu_t;\bphi_t]$};
\path (z.east) ++(0.5,0) coordinate (s0x);
\node[state, anchor=west] (s0) at (s0x |- 0,-0.8) {$\bS_{t-1}$};
\node[op, right=5mm of s0] (dec) {$\Diag(\balpha_t)$};
\node[op, right=5mm of dec] (era) {$\bI-\gamma_t\,\be_t\be_t^{\top}$};
\node[op, right=5mm of era] (del) {$\bI-\beta_t\,\bk_t\bk_t^{\top}$};
\node[op, right=5mm of del] (wr) {$+\,\beta_t\,\bk_t\bv_t^{\top}$};
\node[state, right=5mm of wr] (s1) {$\bS_{t}$};
\node[grey, minimum height=8mm, right=5mm of s1] (rd) {$\bS_t^{\top}\bq_t$};
\node[var, right=4mm of rd] (o) {$\bo_t$};
\foreach \a/\b in {s0/dec, dec/era, era/del, del/wr, wr/s1, s1/rd, rd/o} \draw[arr] (\a) -- (\b);
\foreach \n/\t in {dec/decay, era/erase, del/delta, wr/write, rd/read} \node[note, anchor=base] at (\n.south |- 0,-1.55) {\t};
\draw[arr, black!45] (s1.south) -- ++(0,-0.5) -| (s0.south);
\node[mem] (g) at (z |- 0,0.05) {gate net};
\draw[arr] (z) -- (g);
\draw[barr, -] (g.east) -- (wr.north |- g.east);
\draw[barr] (dec.north |- g.east) -- node[glbl, right, xshift=3pt, pos=0.45] {$\balpha_t$} (dec.north);
\draw[barr] (era.north |- g.east) -- node[glbl, right, xshift=3pt, pos=0.45] {$\gamma_t,\be_t$} (era.north);
\draw[barr] (del.north |- g.east) -- node[glbl, right, xshift=3pt, pos=0.45] {$\beta_t$} (del.north);
\draw[barr] (wr.north |- g.east) -- node[glbl, right, xshift=3pt, pos=0.45] {$\beta_t$} (wr.north);
\end{tikzpicture}
\vspace{-1mm}
\caption{\method{}. Boxes are operations, circles the memory state, bare symbols other variables; blue is the proposed
memory, grey the standard components. (a) Full-history forecaster with seasonal reference $c_{t+h}$ and shrinkage $s_h$.
(b) One erase-then-delta memory step: the gate network reads $\bz_t=[\bu_t;\bphi_t]$ (block input and phase features)
and yields the gates and erase address; $\bq_t,\bk_t,\bv_t$ are projections of $\bu_t$.}
\label{fig:method}
\end{figure*}

\section{Method}
\label{sec:method}
Given the weekly history $\bx_{1:t}\in\Rb^{t\times N}$ of $N$ regions, the target is $\bx_{t+h}$ at lead time $h$.
\method{} (\cref{fig:method}) treats every region as a sequence for temporal mixing and couples regions only
afterwards: per-region normalisation, a linear embedding, $B$ blocks of \{erase-then-delta memory mixer, MLP\}, one
multi-head attention layer across the $N$ regions at every step (optionally with an adjacency matrix as an
attention bias), and a linear head. All components other than the mixer are standard.

\begin{table*}[t]
\centering
\small
\setlength{\tabcolsep}{5pt}
\caption{Results on the three Cola-GNN influenza benchmarks, averaged over lead times $h\in\{3,5,10,15\}$ (pooled RMSE\dn{} / PCC\up{}). Context: history read at forecast time. Reported rows are from~\cite{epignn}, trained rows are averaged over five seeds and the EpiGNN re-run uses the seed median at each lead time. Best bold, second underlined. $^\dagger$Pretraining corpus contains the CDC FluView series of the US datasets (shown, not ranked).}
\label{tab:main}
\vspace{3pt}
\begin{tabular}{llcccccc}
\toprule
Method & Context & \multicolumn{2}{c}{Japan-Pref.} & \multicolumn{2}{c}{US-Regions} & \multicolumn{2}{c}{US-States} \\
 & & RMSE\dn & PCC\up & RMSE\dn & PCC\up & RMSE\dn & PCC\up \\
\rowcolor{gray!15}\multicolumn{8}{l}{\textit{Naive baselines}} \\
Seasonal naive (52 wk) & 1 season & \second{839} & \second{0.913} & 876 & 0.802 & 307 & 0.764 \\
Climatology (2 seasons) & 2 seasons & 1030 & 0.876 & \second{727} & \second{0.862} & 265 & 0.812 \\
\rowcolor{gray!15}\multicolumn{8}{l}{\textit{Epidemic-specific (reported)}} \\
Cola-GNN~\cite{colagnn} & 20 wk & 1254 & 0.839 & 957 & 0.775 & 212 & 0.877 \\
EpiGNN~\cite{epignn} & 20 wk & 1234 & 0.831 & 852 & 0.799 & \second{200} & \second{0.892} \\
EpiGNN~\cite{epignn} (official code, re-run, seed median) & 20 wk & 1379 & 0.765 & 961 & 0.713 & 217 & 0.872 \\
\rowcolor{gray!15}\multicolumn{8}{l}{\textit{General time-series (re-run)}} \\
DLinear~\cite{dlinear} & 20 wk & 1731 & 0.543 & 1013 & 0.695 & 249 & 0.833 \\
PatchTST~\cite{patchtst} & 20 wk & 2026 & 0.185 & 1077 & 0.618 & 251 & 0.818 \\
iTransformer~\cite{itransformer} & 20 wk & 2039 & 0.318 & 991 & 0.687 & 220 & 0.865 \\
TimeMixer~\cite{timemixer} & 20 wk & 1998 & 0.216 & 1109 & 0.599 & 267 & 0.790 \\
TimesNet~\cite{timesnet} & 20 wk & 1965 & 0.284 & 1046 & 0.677 & 288 & 0.776 \\
\rowcolor{gray!15}\multicolumn{8}{l}{\textit{Foundation models (zero-shot)}} \\
Chronos-2~\cite{chronos} (zero-shot) & full & 1314 & 0.776 & 698$^\dagger$ & 0.877$^\dagger$ & 207$^\dagger$ & 0.883$^\dagger$ \\
TimesFM-3~\cite{timesfm} (zero-shot) & full & 1498 & 0.683 & 340$^\dagger$ & 0.972$^\dagger$ & 139$^\dagger$ & 0.949$^\dagger$ \\
\rowcolor{gray!15}\multicolumn{8}{l}{\textit{Ours}} \\
\method{} (window regime) & 20 wk & 1145 & 0.857 & 885 & 0.775 & 206 & 0.885 \\
\method{} (full-history regime) & full & \best{838} & \best{0.926} & \best{698} & \best{0.872} & \best{199} & \best{0.898} \\
\bottomrule
\end{tabular}
\end{table*}

\para{Delta-rule memory for surveillance series.}
The memory has to keep the shape of past seasons while discarding associations that the current phase of the
epidemic has made stale, so we build it from a fast-weight memory with gated decay and an explicitly addressed
erase step whose gates read local phase features.
Linear attention keeps a fast-weight matrix~\cite{schlag2021} and reads it with a query. With $H$ heads of width
$d_h=d/H$, each head holds $\bS_t\in\Rb^{d_h\times d_h}$ and outputs $\bo_t=\bS_t^{\top}\bq_t$. We adopt the
erase-then-delta form~\cite{eda} of the gated delta rule~\cite{deltanet,gdn,kda,gdn2}. For a block input $\bu_t\in\Rb^{d}$ we
form $\bq_t,\bk_t,\bv_t$ by linear maps followed by a short causal convolution and SiLU, with $\bq_t,\bk_t$
$\ell_2$-normalised. The gates have to know where in the wave the region is, so all of them read $\bz_t=[\bu_t;\bphi_t]$, the block
input concatenated with three phase features of the normalised series of the region being mixed, $\bphi_t=[\Delta x_t,\ \Delta^{2}x_t,\
\Delta x_t/(|x_{t-1}|+\epsilon)]$ with $\epsilon{=}0.05$, that is, the growth rate, its curvature and a relative growth rate (clipped to
$\pm5$) that reads the same in small and large regions. The update per head is
\begin{align}
  \tilde{\bS}_t &= \bigl(\bI-\gamma_t\,\be_t\be_t^{\top}\bigr)\,\Diag(\balpha_t)\,\bS_{t-1}, \label{eq:erase}\\
  \bS_t &= \bigl(\bI-\beta_t\,\bk_t\bk_t^{\top}\bigr)\tilde{\bS}_t+\beta_t\,\bk_t\bv_t^{\top}. \label{eq:delta}
\end{align}
\Cref{eq:erase} decays the memory channel-wise and removes the content stored along a learned direction $\be_t$
with strength $\gamma_t=\sigma(\mathbf{w}_\gamma^{\top}\bz_t)$, and \cref{eq:delta} is the gated delta rule with
$\beta_t=\sigma(\mathbf{w}_\beta^{\top}\bz_t)$, which overwrites only the value associated with the write key. The
erase address $\be_t=\mathbf{W}_2\mathbf{W}_1\bz_t/\|\mathbf{W}_2\mathbf{W}_1\bz_t\|_2$ is a factorised map with a 16-dimensional intermediate per head, so stale associations can be cleared along a direction that is decoupled from the key being written.
The decay is $\balpha_t=\exp(-\mathbf{r}\odot\mathbf{m}_t)$ with $\mathbf{m}_t=\mathrm{softplus}(\mathbf{W}_m\bz_t{+}\mathbf{b}_m)$
an input-dependent modulation with learned $\mathbf{W}_m$ and $\mathbf{b}_m$, initialised so that $\mathbf{m}_t=1$, and $\mathbf{r}=\mathrm{softplus}(\boldsymbol{\rho})\in\Rb^{d_h}_{>0}$ learned
per-channel rates whose time constants $\boldsymbol{\tau}=1/\mathbf{r}$ are initialised log-uniformly on $[1,20]$ weeks as in
Mamba~\cite{mamba}, and learning them removes a hand-set hyperparameter. Setting $\gamma_t\equiv0$ recovers Kimi Delta Attention~\cite{kda} (tying the decay across channels as well gives Gated DeltaNet~\cite{gdn}), and replacing \cref{eq:erase} by channel-wise erase/write gates on $\bk_t,\bv_t$ gives Gated DeltaNet-2~\cite{gdn2}. We evaluate both, together with a scalar-decay variant.

\para{Two regimes.}
The benchmark fixes a 20-week input window, whereas a recurrent memory can read an arbitrarily long history, so we
evaluate \method{} in two regimes. Under the standard protocol the model reads a 20-week window and is trained on all windows of the training period
(\emph{window} regime). After early stopping we continue training on the training and validation targets for half
the selected number of epochs at a reduced learning rate (refit). Because the mixer is a recurrence, the same model
can instead read the entire history causally and emit a forecast at every step (\emph{full-history} regime), trained with one gradient step per epoch on the whole training sequence.

\para{Seasonal reference, objective and online adaptation.}
Influenza series repeat with a 52-week period, so in the full-history regime we give the model an explicit seasonal
reference in one of two forms. Either a learned embedding of the week of the year is added to the token embedding,
or the head predicts a correction to a two-season climatology $c_{t+h}=\tfrac{1}{2}\sum_{k=1}^{2}\bar{x}_{t+h-52k}$
($\bar x$: $\pm2$-week mean), $\hat x_{t+h}=c_{t+h}+s_h\,f_\theta(\bx_{1:t})$, with a shrinkage $s_h$ of $0.5$,
$0.3$, $0.1$ and $0.05$ for $h=3$, $5$, $10$ and $15$ because the learned correction is informative at short lead
times and noise at long ones. The training loss is the squared error of the normalised targets, weighted per region
by the square of its scale so that it matches the pooled RMSE of the protocol. Test seasons differ from the training seasons in timing and height, so at test time the model adapts online in one
of two ways. Online refitting takes one gradient step on all observed targets at every new observation,
optionally with Polyak averaging of the weights, and online blending mixes the forecast with the seasonal naive
$x_{t+h-52}$ using the convex weight that minimised the error over the last 12 origins whose targets are known. All
of these operations use only information available at the origin. The components used on each dataset are listed
in \cref{sec:exp}.

\section{Experiments}
\label{sec:exp}
\begin{figure*}[t]
\centering
\includegraphics[width=0.70\textwidth]{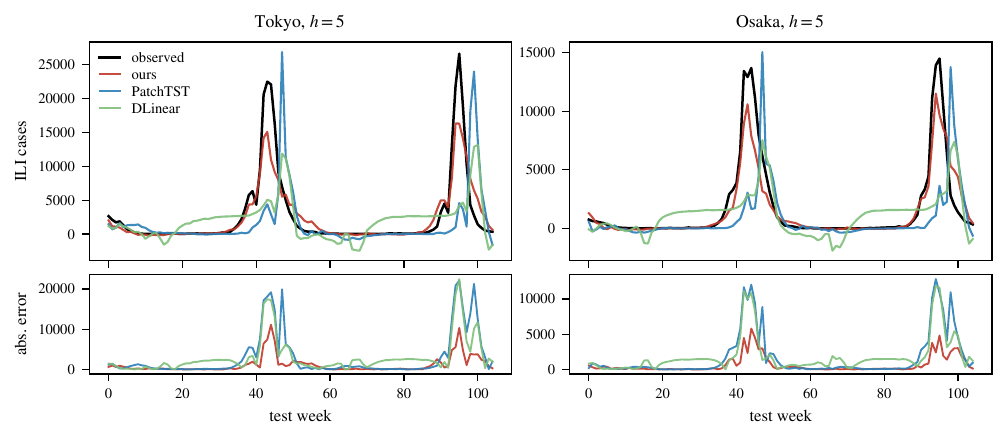}\\[2pt]
\includegraphics[width=0.70\textwidth]{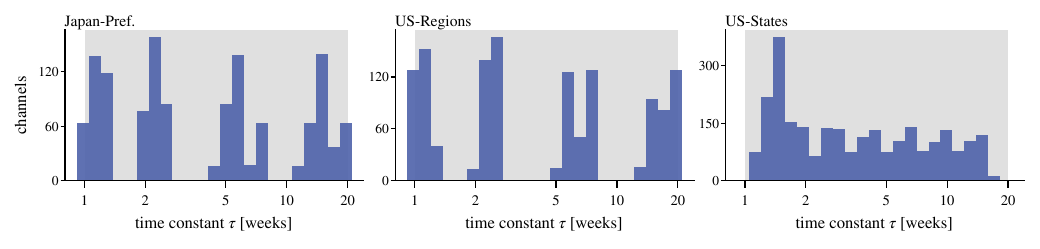}
\vspace{-1mm}
\caption{Top: five-week-ahead forecasts (upper row) and absolute errors (lower row) for Tokyo and Osaka, where PatchTST and
DLinear read the 20-week window. Bottom: learned decay time constants $\tau$ over five seeds and four lead times, with the
initialisation range shaded.}
\label{fig:qual}
\end{figure*}

\para{Epidemic benchmark.}
We follow the Cola-GNN protocol~\cite{colagnn,epignn} on its three influenza datasets: \emph{Japan-Prefectures}
(47 prefectures, 348 weeks, 2012--2019), \emph{US-Regions} (10 HHS regions, 785 weeks, 2002--2017) and
\emph{US-States} (49 states, 360 weeks, 2010--2017). Series are split chronologically 50/20/30 into
train/validation/test, min--max normalised per region with training statistics, and a model receives the last
$L{=}20$ weeks to predict the value $h\in\{3,5,10,15\}$ weeks ahead. We report RMSE and Pearson correlation (PCC) on
the count scale, pooled over all test weeks and regions as in the published tables and averaged over the four lead
times and five seeds. Rows marked \emph{reported} are copied from~\cite{epignn}. All other rows are trained by us
on the same split with the MSE loss, Adam ($10^{-3}$, weight decay $5\times10^{-4}$), batch 128, at most 1500
epochs and early stopping on the validation loss with patience 100, following~\cite{epignn} and its released code.
Besides the epidemic-specific models we re-run five general forecasters from the Time-Series-Library~\cite{tslib}
(DLinear, PatchTST, iTransformer, TimeMixer and TimesNet, with $d_{\text{model}}{=}64$, two layers, the same optimiser
and the same refit as \method{}) and add the zero-shot foundation models Chronos-2~\cite{chronos} and
TimesFM-3~\cite{timesfm} with either the 20-week window or the full history.

\para{Implementation.}
\method{} uses $B{=}2$ blocks, $H{=}8$ heads and a linear head. It uses $d{=}64$ and dropout 0.1 in the window regime (125k parameters), $d{=}32$ and dropout 0.5 in the full-history regime (40k parameters), except on US-States ($d{=}64$, dropout 0.4). The memory, embedding and head are identical across datasets, whereas the seasonal and online
components differ. Japan uses the week-of-year embedding, the scale-weighted loss and online blending, US-Regions
the climatology correction with $s_h$, the scale-weighted loss and online refitting at $h\le5$, US-States the
adjacency bias, Polyak averaging and online refitting. Beyond these listed choices nothing differs between datasets or lead times, and \cref{tab:ablation} quantifies every component on every dataset where it is used. A full-history run takes about 40 minutes on one NVIDIA RTX PRO 6000 GPU, whereas window baselines take seconds.

\para{Findings about the benchmark.}
Three facts not reported before frame the comparison. (i) \emph{Seasonality dominates, but validation cannot see
it.} The seasonal naive forecast $x_{t+h-52}$ outperforms every published GNN at $h{=}10$ and $15$ on Japan and
US-Regions (\cref{tab:main}), yet it is worse on the validation than on the test seasons (Japan RMSE 1862 vs.\ 839), so no validation-based procedure selects it. (ii) \emph{Zero-shot foundation models are contaminated on the
US datasets.} The GiftEvalPretrain corpus~\cite{gifteval} used by TimesFM-3 and Chronos-2 contains the CDC FluView ILINet and WHO/NREVSS series underlying US-Regions and US-States. The RMSE of TimesFM-3 on
US-Regions barely depends on the lead time (318--355), unlike on the Japanese data, which is not in the corpus. (iii) \emph{Their strength is context length.} With the 20-week window both are worse than DLinear on all three datasets.

\begin{table}[t]
\centering
\small
\setlength{\tabcolsep}{2pt}
\caption{Ablation study in the full-history regime, removing or replacing one component at a time (seeds 0--2, averaged over the four lead times). Bold: best per column. Dashes mark components not used on a dataset.}
\label{tab:ablation}
\vspace{3pt}
\begin{tabular}{lcccccc}
\toprule
Variant & \multicolumn{2}{c}{Japan-Pref.} & \multicolumn{2}{c}{US-Regions} & \multicolumn{2}{c}{US-States} \\
 & RMSE\dn & PCC\up & RMSE\dn & PCC\up & RMSE\dn & PCC\up \\
\midrule
GDN-2 rule & 871 & 0.919 & 707 & 0.868 & 202 & 0.896 \\
no erase (KDA) & 860 & 0.923 & 712 & 0.867 & 208 & 0.890 \\
no decay & 859 & 0.916 & 728 & 0.869 & 250 & 0.864 \\
no phase feat. & 880 & 0.916 & \best{690} & 0.875 & 201 & 0.896 \\
no region attn. & 865 & 0.924 & 691 & 0.875 & 213 & 0.880 \\
softmax mixer & 874 & 0.913 & 713 & 0.866 & 238 & 0.846 \\
no seasonal ref. & 865 & 0.917 & 941 & 0.737 & -- & -- \\
constant $s_h$ & -- & -- & 717 & 0.862 & -- & -- \\
unweighted loss & 855 & 0.926 & 734 & 0.858 & -- & -- \\
no online blend & 1063 & 0.892 & -- & -- & -- & -- \\
no online refit & -- & -- & 698 & 0.874 & 199 & 0.893 \\
no Polyak avg. & -- & -- & -- & -- & 196 & \best{0.900} \\
no adjacency bias & -- & -- & -- & -- & 197 & 0.898 \\
\midrule
\method{} (full) & \best{833} & \best{0.927} & 691 & \best{0.876} & \best{195} & 0.899 \\
\bottomrule
\end{tabular}
\end{table}

\para{Results on the epidemic benchmark.}
In the window regime (\cref{tab:main}) \method{} is the best trained model on Japan (RMSE 1145 vs.\ 1234 for the
reported EpiGNN) and behind it on the two US datasets (885 vs.\ 852, 206 vs.\ 200). Re-running the official EpiGNN code with five seeds gives 1379,
961 and 217 on Japan, US-Regions and US-States (seed median at each lead time, as some seeds diverge). Against these our margins are 17\%, 8\% and 5\%. The full-history regime lowers
\method{}'s RMSE by 27\%, 21\% and 3\% relative to the window regime and is the best uncontaminated entry on all
three datasets. It ties the seasonal naive on Japan in RMSE (838 vs.\ 839) and exceeds it in PCC (0.926 vs.\
0.913). No published, re-run or zero-shot model comes within 47\% of that naive. It outperforms the climatology on US-Regions (698 vs.\ 727) and equals the contaminated Chronos-2 there, and it
matches the reported EpiGNN on US-States (199 vs.\ 200). The gain concentrates at short lead times. \method{} is
20\% below the climatology at $h{=}3$ on US-Regions but 5.5\% above it at $h{=}10$, and 2.9\% above the seasonal
naive at $h{=}5$ on Japan. Per region, its RMSE is below that of the seasonal naive in 64--77\% of prefectures at every lead
time and the climatology in 67--96\% of US states, but only at $h{=}3$ (10/10) and $h{=}15$ (6/10) on the HHS
regions.

Part of the full-history gain is context rather than memory. Giving the five general forecasters the same
post-hoc online blend lifts the best of them to 864, 697 and 220 (not shown), which is level with \method{} on
US-Regions, 3\% behind on Japan and 10\% behind on US-States, and the softmax-mixer row of \cref{tab:ablation}
isolates the memory.

\para{Ablation study.}
\Cref{tab:ablation} removes or replaces one component at a time. Every row uses seeds 0--2, where the full model
reaches RMSE 833, 691 and 195. (i) \emph{The memory rule matters.} Erase-then-delta is the best rule on all three datasets. Kimi Delta Attention and Gated DeltaNet-2 follow (860--871, 707--712, 202--208), and a softmax-attention mixer costs 5\%, 3\% and 22\% (874, 713, 238). Removing the decay raises RMSE to 728 and 250 on the US datasets. (ii) \emph{Phase features and region attention help where
regions are heterogeneous.} Removing the phase features raises RMSE to 880 on Japan and 201 on US-States, and
removing the region attention to 865 and 213, whereas the ten HHS regions stay at 690 and 691. The gates recorded over the test period barely move (erase gate near 0.2, write gate near 0.5 in all 20
Japan runs), so the phase features act through the erase address $\be_t$ rather than through the timing of the gates. (iii) \emph{The
seasonal components dominate where they are used.} Without the seasonal reference Japan loses 4\% (865) and
US-Regions breaks down (941), without online blending Japan worsens to 1063, a constant shrinkage costs US-Regions
27 RMSE (717), and the unweighted objective costs US-Regions 43 (734) and Japan 22 (855). Online refitting, Polyak averaging and the adjacency bias change RMSE by four points at most.

\para{Qualitative analysis.}
\Cref{fig:qual} (top) shows five-week-ahead forecasts for Tokyo and Osaka over the two test seasons. \method{} places
all four peaks within one week of the observed peak, whereas PatchTST and DLinear peak 3--5 weeks late. The bottom row shows the learned time constants $\tau$. A memory
that learned the annual cycle itself would drive some of them towards one season, yet all stay inside the 1--20 week
initialisation range (median 3--4 weeks). Widening the initialisation to $[1,104]$ weeks lets 12--25\% of the channels
settle beyond one season, but RMSE changes to 857, 697 and 202 against 838, 698 and 199, so the annual cycle is better
supplied by the explicit seasonal reference.

\section{Conclusion}
\label{sec:conclusion}
We introduced \method{}, a forecaster built around a delta-rule memory with phase-conditioned erase and write, an explicit seasonal reference and online adaptation. On the three Cola-GNN benchmarks it is the best uncontaminated model, and with everything else held fixed the memory contributes 3--22\% RMSE over softmax attention. A seasonal naive forecast outperforms every published graph model at long lead times on two datasets. The main limitation is the scope of the evaluation, which is confined to influenza surveillance on three benchmarks. In future work, we plan to add probabilistic outputs and to cover other diseases and hospitalisation targets.


\bibliographystyle{IEEEbib}
\bibliography{refs}

\begin{thebibliography}{10}

\bibitem{mathis2024}
Sarabeth~M. Mathis et~al.,
\newblock ``Evaluation of {FluSight} influenza forecasting in the 2021--22 and
  2022--23 seasons with a new target laboratory-confirmed influenza
  hospitalizations,''
\newblock {\em Nature Communications}, vol. 15, no. 1, 2024,
\newblock Art. no. 6289.

\bibitem{colagnn}
Songgaojun Deng, Shusen Wang, Huzefa Rangwala, Lijing Wang, and Yue Ning,
\newblock ``{Cola-GNN}: Cross-location attention based graph neural networks
  for long-term {ILI} prediction,''
\newblock in {\em Proc. ACM CIKM}, 2020, pp. 245--254.

\bibitem{epignn}
Feng Xie, Zhong Zhang, Liang Li, Bin Zhou, and Yusong Tan,
\newblock ``{EpiGNN}: Exploring spatial transmission with graph neural network
  for regional epidemic forecasting,''
\newblock in {\em Proc. ECML PKDD 2022, LNCS}, 2023, pp. 469--485.

\bibitem{einns}
Alexander Rodr{\'\i}guez et~al.,
\newblock ``{EINNs}: Epidemiologically-informed neural networks,''
\newblock in {\em Proc. AAAI Conf. Artif. Intell.}, 2023, vol.~37, pp.
  14453--14460.

\bibitem{camul}
Harshavardhan Kamarthi et~al.,
\newblock ``{CAMul}: Calibrated and accurate multi-view time-series
  forecasting,''
\newblock in {\em Proc. ACM Web Conf. (WWW)}, 2022, pp. 3174--3185.

\bibitem{patchtst}
Yuqi Nie, Nam~H. Nguyen, Phanwadee Sinthong, and Jayant Kalagnanam,
\newblock ``A time series is worth 64 words: Long-term forecasting with
  transformers,''
\newblock in {\em Proc. ICLR}, 2023.

\bibitem{itransformer}
Yong Liu et~al.,
\newblock ``{iTransformer}: Inverted transformers are effective for time series
  forecasting,''
\newblock in {\em Proc. ICLR}, 2024.

\bibitem{timemixer}
Shiyu Wang et~al.,
\newblock ``{TimeMixer}: Decomposable multiscale mixing for time series
  forecasting,''
\newblock in {\em Proc. ICLR}, 2024.

\bibitem{dlinear}
Ailing Zeng et~al.,
\newblock ``Are transformers effective for time series forecasting?,''
\newblock in {\em Proc. AAAI}, 2023, vol.~37, pp. 11121--11128.

\bibitem{deltanet}
Songlin Yang, Bailin Wang, Yu~Zhang, Yikang Shen, and Yoon Kim,
\newblock ``Parallelizing linear transformers with the delta rule over sequence
  length,''
\newblock in {\em Adv. Neural Inf. Process. Syst. (NeurIPS)}, 2024, vol.~37,
  pp. 115491--115522.

\bibitem{gdn}
Songlin Yang et~al.,
\newblock ``Gated delta networks: Improving {Mamba2} with delta rule,''
\newblock in {\em Proc. ICLR}, 2025.

\bibitem{kda}
{Kimi Team},
\newblock ``{Kimi Linear}: An expressive, efficient attention architecture,''
\newblock {\em arXiv preprint arXiv:2510.26692}, 2025.

\bibitem{gdn2}
Ali Hatamizadeh et~al.,
\newblock ``Gated {{\mbox{DeltaNet-2}}}: {{\mbox{Decoupling}}} erase and write
  in linear attention,''
\newblock {\em arXiv preprint arXiv:2605.22791}, 2026.

\bibitem{eda}
Xiao Li et~al.,
\newblock ``Erase-then-delta attention: Decoupling erase and write addresses in
  delta-rule linear attention,''
\newblock {\em arXiv preprint arXiv:2606.26560}, 2026.

\bibitem{spatialepibench}
Ruiqi Lyu et~al.,
\newblock ``{\mbox{SpatialEpiBench}}: Benchmarking spatial information and
  epidemic priors in forecasting,''
\newblock {\em arXiv preprint arXiv:2605.06530}, 2026.

\bibitem{epicastbench}
Madhurima Panja, Danny D'Agostino, Huitao Li, Tanujit Chakraborty, and Nan Liu,
\newblock ``{\mbox{EpiCastBench}}: Datasets and benchmarks for multivariate
  epidemic forecasting,''
\newblock {\em arXiv preprint arXiv:2605.11598}, 2026.

\bibitem{timesnet}
Haixu Wu et~al.,
\newblock ``{TimesNet}: Temporal {2D}-variation modeling for general time
  series analysis,''
\newblock in {\em Proc. ICLR}, 2023.

\bibitem{decompssm}
Shunya Nagashima, Shuntaro Suzuki, Shuitsu Koyama, and Shinnosuke Hirano,
\newblock ``A decomposition-based state space model for multivariate
  time-series forecasting,''
\newblock in {\em Proc. IEEE Int. Conf. Acoustics, Speech and Signal Processing
  (ICASSP)}, 2026, pp. 1241--1245.

\bibitem{chronos}
Abdul~Fatir Ansari et~al.,
\newblock ``{Chronos-2}: From univariate to universal forecasting,''
\newblock {\em arXiv preprint arXiv:2510.15821}, 2025.

\bibitem{timesfm}
Abhimanyu Das, Weihao Kong, Rajat Sen, and Yichen Zhou,
\newblock ``A decoder-only foundation model for time-series
  {\mbox{forecasting}},''
\newblock in {\em Proc. ICML, PMLR~235}, 2024, pp. 10148--10167.

\bibitem{schlag2021}
Imanol Schlag, Kazuki Irie, and J{\"u}rgen Schmidhuber,
\newblock ``Linear transformers are secretly fast weight programmers,''
\newblock in {\em Proc. ICML, PMLR~139}, 2021, pp. 9355--9366.

\bibitem{mamba}
Albert Gu and Tri Dao,
\newblock ``Mamba: Linear-time sequence {\mbox{modeling}} with selective state
  spaces,''
\newblock in {\em Proc. COLM}, 2024.

\bibitem{tslib}
Yuxuan Wang et~al.,
\newblock ``Deep time series models: A {\mbox{comprehensive}} survey and
  benchmark,''
\newblock {\em IEEE Trans. Pattern Analysis and Machine Intelligence}, 2026,
\newblock early access, doi: 10.1109/TPAMI.2026.3690845.

\bibitem{gifteval}
Taha Aksu et~al.,
\newblock ``{\mbox{GIFT-Eval}}: A benchmark for general time series forecasting
  model evaluation,''
\newblock {\em arXiv preprint arXiv:2410.10393}, 2024.

\end{thebibliography}
\end{document}